%% file: main.tex
\documentclass[runningheads]{llncs}

\usepackage{eccv}

\usepackage{eccvabbrv}

\usepackage{graphicx}
\usepackage{wrapfig}  %
\usepackage{newunicodechar}
\newunicodechar{∼}{\textasciitilde{}}
\usepackage{booktabs}
\usepackage{microtype}
\usepackage{multirow}  %
\usepackage{sourcesanspro}  %

\usepackage[accsupp]{axessibility}  %

\usepackage[pagebackref,breaklinks,colorlinks,linkcolor=blue,citecolor=eccvblue,filecolor=magenta,urlcolor=blue]{hyperref}

\hypersetup{bookmarksdepth=2}

\newcommand{\ours}{\emph{Pictura}\xspace}
\newcommand{\agent}{\emph{Alberti}\xspace}

\begin{document}

\title{\ours: Perspective-View Self-Play \texorpdfstring{\\}{}at Scale for Driving}

\titlerunning{\ours: Perspective Self-Play for Driving}

\author{Yuan Yin\inst{1} \and
Elias Ramzi\inst{1} \and
Marc Lafon\inst{1} \and
Valentin Charraut\inst{2} \and
Victor Bares\inst{2}\and Yihong Xu\inst{1} \and Éloi Zablocki\inst{1} \and Alexandre Boulch\inst{1} \and Thibault Buhet\inst{2} \and Andrei Bursuc\inst{1} \and Matthieu Cord\inst{1,3}}

\authorrunning{Y.~Yin et al.}

\institute{valeo.ai, Paris, France \and
Valeo Brain, Créteil, France \and
Sorbonne Université, CNRS, ISIR, F-75005 Paris, France}

\maketitle

\begin{abstract}
Self-play in simulation produces robust driving policies at scale.
Demonstrations of such behavior have been made using \textit{privileged} vectorized observations such as exact poses and velocities, even for occluded agents.
This assumes that perception is solved and introduces a representation gap with the partial observation of a deployed agent driving from the perspective view of egocentric cameras.
A common fix, distilling the privileged policy into a camera-input student, leaves the student imitating decisions its own view cannot justify.
Instead, we establish perspective-view self-play as a practical training regime.
We introduce \ours, a GPU-accelerated multi-agent driving simulator that renders each agent's egocentric view at every step, mitigating the representation gap at its source.
\ours sustains up to 500\,K agent-steps/s (2\,M images/s) on a single H100.
Using \ours, we train \agent by self-play with plain PPO.
It is the first large-scale driving self-play policy trained directly from perspective images, without privileged observations.
Training spans 50\,B agent steps for $\sim$35\,M km of driving. 
It approaches the driving performance of its privileged vectorized counterpart, and transfers zero-shot to Waymo Open Motion Dataset layouts re-rendered in \ours, where it outperforms privileged vectorized agents.
Project page: \url{https://valeoai.github.io/Pictura/}.
\end{abstract}

\input{sections/01_introduction}

\input{sections/02_related_work}
\input{sections/03_method}

\input{sections/04_rasterizer_eval}

\input{sections/05_policy_eval}

\input{sections/06_conclusion}

\section*{Acknowledgments}
We thank Ellington Kirby, whose earlier work on the vector simulator laid the groundwork for this project; Waël Doulazmi and the rest of the Valeo DRIL team for many valuable discussions on simulation and reinforcement learning; and Lan Feng for generous discussions and exploratory work on a related direction. This project was provided with computing AI and storage resources by GENCI at IDRIS thanks to the grants A0201016239, A0181016239 and A0181016203 on the supercomputer Jean Zay's H100 partition. We acknowledge EuroHPC Joint Undertaking for awarding the project IDs EHPC-AIF-2026FL01-075 and EHPC-AIF-2026FL01-258 access to MareNostrum5 at BSC, Spain.
This work was supported by the VISA DEEP AI chair, funded by the Agence Nationale de la Recherche (ANR, ANR-20-CHIA-0022), and by the PostGenAI@Paris cluster, which benefited from government funding managed by the ANR under the France 2030 program (ANR-23-IACL-0007).

\bibliographystyle{splncs04}
\bibliography{main}

\clearpage

\input{sections/0X_appendix}

\end{document}

%% file: sections/01_introduction.tex
\section{Introduction}
\label{sec:intro}

\input{figs/teaser}

End-to-end driving policies are usually learned by imitating logged human driving~\cite{pomerleau1988alvinn,uniad,drivor}.
Offline imitation has fundamental limits.
It demands large amounts of real driving data, which only a few established industrial players can collect.
And it exposes the policy only to the states a careful human visited, so covariate shift leaves it unable to recover from its own mistakes.
Multi-agent reinforcement learning (MARL) through self-play removes these limits by having agents train against copies of themselves in simulation, generating ever-harder scenarios and learning from the consequences of their actions.
Recent work shows robust driving behavior emerges from self-play at scale, with no human demonstrations, transferring zero-shot to maps and agents unseen in training~\cite{gigaflow}.

This robustness, however, emerges only after billions of simulator interactions, so self-play is practical only with fast, batched simulation~\cite{gigaflow,pufferdrive}.
Part of this speed comes from what the agents observe: a vectorized state, a compact list of nearby agents, lanes, and traffic signals, copied directly from the internal simulation state~\cite{gigaflow,gpudrive,pufferdrive}.
This state is privileged, handing the policy information that no on-board sensor can measure, such as occluded agents or the exact velocities of partners.
A deployed vehicle instead perceives the world through egocentric cameras and must infer the scene geometry, dynamics, and semantics from raw pixels.
Training on privileged vectors thus solves an easier problem than deployment poses.
Worse, such a policy grounds its decisions in information a camera cannot supply, reacting to agents and signals it could never actually perceive.
An end-to-end policy distilled from it~\cite{gigapixel,terratransfer} therefore inherits decisions it cannot justify from its own view, a failure mode LEAD~\cite{lead} documents and our analysis corroborates.

Existing large-scale self-play learns its policy from privileged simulator state and distills it to images only afterward. We show that this privileged stage is unnecessary: reinforcement learning can be performed directly from observable images, with the policy learning to drive from the \emph{perspective} camera view during self-play itself.
This enforces an observability constraint the vectorized state violates. The perspective view is partial: it hides occluded objects, reduces distant ones to a few pixels, and carries no metric description of the scene.
Training under this view therefore prevents the policy from exploiting privileged scene information: it sees nothing that a deployed vehicle could not perceive for itself.
Photorealistic images would satisfy that constraint most faithfully, but rendering them in closed loop at self-play scale is prohibitively expensive. We take a middle ground that keeps the observability constraint and drops only the appearance: a rasterized perspective view of the simulator state, cheap to render at scale.
This costs less than it might seem. Growing evidence indicates that scene structure and dynamics, more than appearance, carry the learning signal, and that models trained on cheap non-photorealistic sources transfer to real images with limited real data, e.g., in generation~\cite{DBLP:journals/corr/abs-2601-09452,DBLP:conf/nips/YangCGCSJLGYC25}, in planning~\cite{rap,gigapixel}.
Aligning the rendered view with real imagery is a complementary, orthogonal step, left to future work.

We introduce \ours\footnote{\ours and \agent are named after Leon Battista Alberti's \emph{De Pictura} (1435), the treatise that first formalized the mathematical rules of linear perspective.}, a GPU-accelerated multi-agent driving simulator that makes perspective self-play practical at scale (\autoref{fig:teaser}).
At every simulation step, it renders the rasterized egocentric view of every agent and feeds it back to that agent's policy --- each agent acts, is rendered for its neighbors, and learns from what it sees, all within one closed loop.
Perception and control are learned jointly, without a privileged teacher or distillation procedure; the resulting policy, \agent, thus learns to drive from images within the self-play loop itself, which to our knowledge no prior driving policy does.
Trained only on what a camera sees, \agent develops reasonable, human-like behavior, slowing through occlusion-heavy scenes and where its sight ends, and re-accelerating once confident the road ahead is clear.
The whole loop runs on synthetic maps imported from CARLA~\cite{carla}.
Trained this way, \agent transfers zero-shot to real-world layouts: road maps, agent initial states, and goals drawn from the Waymo Open Motion Dataset (WOMD)~\cite{womd} logged driving, re-rendered through \ours's rasterizer.

Our contributions are:

\noindent(1) \textbf{self-play from on-board observations at scale}: we show that large-scale driving self-play, until now confined to privileged vector states, can run directly on perspective images, removing privileged information from the reinforcement-learning stage entirely;

\noindent(2) \textbf{\ours, the simulator with a GPU-native renderer}:
a complete vectorized reimplementation of Gigaflow's world design, enriched beyond moving-vehicle boxes with traffic lights, pedestrians, cyclists, walls, and parked cars; its custom CUDA rasterizer draws every agent's camera views inside the training loop at 500\,K agent-steps per second (SPS; 2\,M images/s) on a single H100;%

\noindent(3) \textbf{\agent, the resulting policy}: trained from scratch with plain PPO on rendered camera views, without privileged teacher or distillation. It drives nearly as well as its privileged vectorized counterpart while transferring better zero-shot to real-world WOMD layouts.

%% file: figs/teaser.tex
\begin{figure}[t]
  \centering
  \includegraphics{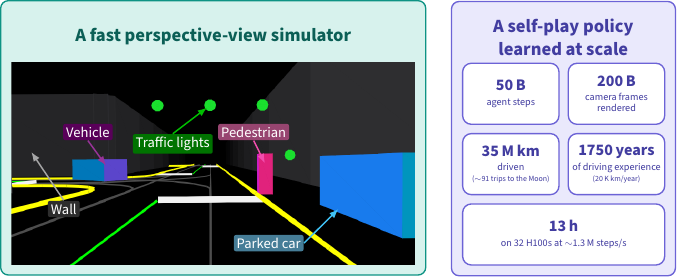}
\caption{\textbf{\ours runs driving self-play directly in perspective view.}
Left: a rendered frame of the kind the policy learns from, produced by our
custom rasterizer; the labels mark the scene elements it supports beyond moving vehicles.
Right: the scale of the longest self-play run of \agent.}
  \label{fig:teaser}
\end{figure}

%% file: sections/02_related_work.tex
\section{Related Work}
\label{sec:related}

\paragraph{Self-play for driving.}
Self-play is a classic recipe for multi-agent RL~\cite{lanctot2017unified,silver2018general,tesauro1995temporal}.
In driving it yields negotiation behaviors such as merging in simplified
environments~\cite{DBLP:conf/iccvw/Tang19}; regularizing it with human data keeps agents
compatible with human conventions~\cite{DBLP:journals/corr/abs-2403-19648,corneliss2026pinch}; and asymmetric
variants train against a teacher proposing challenging yet solvable scenarios, improving
long-tail robustness~\cite{DBLP:conf/eccv/ZhangBWFZCCU24}. Gigaflow~\cite{gigaflow} shows that at sufficient scale, robust behavior emerges from self-play alone: with no real-world data or annotations, its policy reaches state-of-the-art results on CARLA~\cite{carla}, nuPlan~\cite{nuplan} and WOMD~\cite{womd}, and transfers zero-shot to unseen maps. This comes at a price. The regime is only reachable in a simulator that steps billions of transitions, and every agent in it accordingly acts on a vectorized privileged state, which very few to no on-board sensors can supply.

\paragraph{Closed-loop driving simulators.}
That step rate is what dictates the observation.
Vectorized simulators~\cite{pufferdrive,gpudrive,waymax} step abstract state at millions of transitions per second and are the standard substrate for RL, but expose poses, velocities and road-graph points rather than anything a camera could produce.
GPUDrive~\cite{gpudrive} is representative. It builds on the Madrona batch engine~\cite{madrona}, uses it for state stepping only, and reports the throughput of a simulator that renders nothing.
Perspective simulators make the opposite trade, rendering images from authored assets and a game engine~\cite{li2023metadrive,carla} or from per-scene 3D reconstruction~\cite{amini2022vista2,ljungbergh2024neuroncap,hugsim}, and neither approaches the environment counts self-play requires.
No existing driving simulator delivers a camera-like observation at vectorized throughput.

\paragraph{From privileged agents to camera-input students.}
Fast simulators are blind and sighted ones are slow, so the field trains in the fast one and transfers the policy to camera input.
A privileged agent supervises a vision-based student~\cite{DBLP:conf/corl/0001ZKK19,DBLP:conf/iccv/ZhangLDYG21,DBLP:conf/iccv/0001KK21}, or is frozen behind a learned adapter~\cite{DBLP:conf/iccv/JiaGCYLL23}.
Recently, TerraTransfer~\cite{terratransfer} aligns a vision backbone with the latent space of a vectorized RL agent; Gigapixel~\cite{gigapixel} distills a reactive agent into an imitation-learning planner~\cite{drivor} through an intermediate pixel observation.
Either way, the student is trained to reproduce decisions that depend on what it cannot see~\cite{lead}.
\ours removes the premise rather than improving the transfer: once the perspective observation is cheap enough to sit inside the RL loop, the camera-input policy can be trained directly.

\paragraph{Perspective rasterization of abstract state.}
Such an observation is obtainable by rasterizing the simulator state directly into the perspective
view, dropping appearance entirely. Prior work uses such rasters only to condition generative
models~\cite{li2023drivingdiffusion,wen2023panacea,nvidia2025cosmosdrivedreams}. RAP~\cite{rap} is
the first to use the raster to \emph{enhance} a perspective-view driving agent, through behavioral
augmentation for imitation learning, but rasterizes on the host with OpenCV, which is adequate only offline. Gigapixel~\cite{gigapixel} renders with Madrona's batch renderer~\cite{madrona},
which routes through the GPU's graphics pipeline, whose capacity is reduced on modern, computation-oriented hardware. 
Scaling self-play efficiently therefore favors producing the observation on the same device and in the same process as the policy. \ours does this by rasterizing on the GPU's general-purpose compute cores rather than its graphics pipeline. A driving scene, a handful of boxes, lanes, and polylines, is simple enough to render this way.

%% file: sections/03_method.tex
\section{\ours: A Simulator for Perspective-View Self-Play}
\label{sec:method}

\ours is a GPU-accelerated simulator for multi-agent driving self-play in perspective view. It has three parts: the vectorized simulator at its core (\autoref{sec:method:sim}), the perspective rasterizer (\autoref{sec:method:render}), and the training recipe and policy behind \agent (\autoref{sec:method:training}), shown together in \autoref{fig:system}.

\input{figs/system}

\subsection{A Gigaflow-Style Vectorized Simulator}
\label{sec:method:sim}

Self-play yields robust driving only at scale~\cite{gigaflow}. Reaching that scale places two demands on the simulator: it must step transitions cheaply enough to accumulate billions of them, and its world---heavy randomization, carefully shaped rewards---must be varied enough to force generalization. This section describes the vectorized core that meets both; the perspective renderer (\autoref{sec:method:render}) and the image-based policy (\autoref{sec:method:training}) are layered on top of it.

\paragraph{Building on PufferDrive.}
We build on PufferDrive~\cite{pufferdrive}, an open-source driving simulator written with PufferLib~\cite{pufferlib} and, to our knowledge, the best public starting point for vectorized self-play RL in driving. From it, \ours inherits a fast C multi-agent driving core, a PyTorch PPO training loop, and support for over a thousand agents per environment.
All scenarios run on the synthetic CARLA~\cite{carla} maps PufferDrive distributes, whose road networks are hand-authored, realistic, and diverse.
Episodes are open-ended: on reaching its goal, an agent is immediately assigned a new one and keeps driving.
As released, though, PufferDrive lacks several of Gigaflow's features~\cite{gigaflow}; the next paragraphs describe the changes that add them, turning the core into a Gigaflow-style simulator.

\paragraph{Agents, scene, and goals.}
A single shared policy controls every agent on the road, which lets \ours populate the scene more richly. Pedestrians and cyclists join the vehicles as controllable agents. Parked cars are added as static entries, and walls become map elements that later act as occluders for the renderer. Two further changes shape the driving task. Goals are sampled along the lane network rather than anywhere in the drivable area, so reaching one requires lane-consistent driving. Infractions are terminal: an agent that collides, leaves the road, or runs a red light halts in place for the rest of the episode. The episode resets once the halted fraction exceeds a threshold (\autoref{sec:appendix:training:opt}, \autoref{tab:hparams}).

\paragraph{Observation, reward, and conditioning.}
At every step $t$, agent $i$ receives an egocentric, normalized observation built from the ego vehicle state $S$, the nearest surrounding agents $A$ (position, orientation, velocity, and size, parked cars included), and a map set $\mathcal{W}_i^{(t)}$ that gathers the nearby lane geometry $W_{\text{lane}}$, the road boundaries $W_{\text{boundary}}$, the traffic-control elements $W_{\text{stop}}$ (stop lines and traffic-light states), and, when enabled, the walls $W_{\text{wall}}$:
\begin{equation}
  o_i^{(t)} = \big(S_i^{(t)}, A_i^{(t)}, \mathcal{W}_i^{(t)}, \mathcal{C}_i\big),
  \;\;
  \mathcal{W}_i^{(t)} = \big\{W_{\text{lane},i}^{(t)}, W_{\text{boundary},i}^{(t)}, W_{\text{stop},i}^{(t)}, W_{\text{wall},i}^{(t)}\big\}.
  \label{eq:obs_vec}
\end{equation}
The observation also carries a per-agent conditioning set $\mathcal{C}_i = (C_{\text{reward}}, C_{\text{dynamics}})$, which tells each agent the reward weighting and dynamics it is currently subject to.
Each agent maximizes its own per-step reward, a sum of terms that reward waypoint and goal attainment and penalize collisions, leaving the road, crossing a stop line on red, and other infractions. At every episode reset, most reward coefficients are resampled, along with the dynamics multipliers $C_{\text{dynamics}}$ that scale each agent's controls.
Because $\mathcal{C}_i$ is part of the observation, the shared policy can adapt its behavior to each draw.
\autoref{sec:appendix:reward} defines all reward terms, and \autoref{tab:reward-dist} lists the sampling distribution of every conditioning coefficient.

\input{figs/qualitative}

\subsection{Perspective Rendering}
\label{sec:method:render}

Perspective self-play replaces the vectorized observation of \autoref{sec:method:sim} with each agent's egocentric camera view. Rendering it for every agent at every step must not come to dominate training cost. \ours meets this with a purpose-built batch rasterizer whose visual design follows RAP~\cite{rap}: the scene is drawn as flat-shaded geometric primitives, without textures or lighting, a lightweight representation that still carries the cues a driving policy needs.

\paragraph{Rasterizer.}
The \ours rasterizer is a deterministic function $\Phi$: for each agent $i$ it projects the surrounding agents and the map geometry onto a rig $\mathcal{K}_i$ of $N_{\text{cam}}$ ego-mounted cameras with per-camera configured intrinsics and extrinsics, producing the perspective observation the policy consumes,
\begin{equation}
  \hat{o}_i^{(t)}\!=\!\big(S_i^{(t)},\!I_i^{(t)},\!\mathcal{C}_i\big),
 I_i^{(t)}\!=\!\big( I_{i,k}^{(t)} \big)_{k=1}^{N_{\text{cam}}}, I_{i,k}^{(t)} = \Phi\big(A_i^{(t)},\!\mathcal{W}_i^{(t)},\!\mathcal{K}_{i,k}\big)\!\in\![0,1]^{h_k\!\times\!w_k\!\times\!3}.
\label{eq:obs_pv}
\end{equation}
Each image is obtained by rasterizing the scene primitives---agent cuboids, building faces, and map polylines---under the ego camera projection. Only the surroundings are rendered: the ego state  $S_i$ and the conditioning $\mathcal{C}_i$ stay vector inputs, matching what is observable on-board at test time.

\paragraph{Rendering features for driving.}
\autoref{fig:qualitative} shows the scene elements \ours draws into the rendered view.
The surrounding agents are cuboids colored per face with depth-aware brightness, so their orientation and range read from color alone: pedestrians, cyclists, and vehicles, each in its own hue~(a--c).
Traffic lights render as disks that are visible only within a limited angular range. A signal is drawn for the approach it governs~(d) yet culled once the ego faces away~(e), matching how a real signal serves a single approach. The signal's position is also decoupled from the stop line it controls. As in American-style intersections, the head can hang on the far side of the junction, while governing the near-side stop line~(f).
Static geometry completes the view: parked cars line the curb as uncontrolled agents~(g), the walls $W_{\text{wall}}$ render as occluders that model the limited visibility of city streets~(h), and road markings are polylines whose width shrinks with depth, so nearer markings read as thicker~(i).
Underpinning all of these, analytic-coverage antialiasing smooths edges at a single sample per pixel (far cheaper than supersampling or multisampling), so thin structures such as lane lines remain visible even at low resolution (\autoref{sec:appendix:render-quality}, \autoref{fig:town_scenes}).
Training can therefore run at low resolution, reaching the billions of agent steps self-play demands (\autoref{fig:resolution}).%

\paragraph{A renderer built for the training stack.}
Because the scene contains only basic shapes, \ours rasterizes it with a custom CUDA kernel on the GPU's general-purpose compute cores, the same cores that run the network, rather than the fixed-function graphics pipeline a conventional renderer drives. Three consequences follow: (1) Rendered views feed the network on-device with no host round-trip, because the kernel runs inside the training process, unlike a separate renderer that pays that round-trip every step. (2) The renderer speeds up with each new hardware generation, because general-purpose compute is what deep-learning hardware keeps accelerating, while the graphics pipeline is deprioritized. (3) It runs wherever the training stack does, including shared HPC clusters that lack a full graphics stack, because the kernel compiles just-in-time for any CUDA-capable device, needing no graphics driver or prebuilt binary.

\subsection{Perspective-View Self-Play}
\label{sec:method:training}
Coupling the vectorized simulator with the rasterizer, the agent \agent learns directly from images, using reinforcement learning as its only supervision.
It is trained from scratch by multi-agent self-play PPO on the rendered views, jointly optimizing the image encoder and the driving policy.

\paragraph{Learning objective.}
The perspective agent optimizes the same objective as the vectorized baseline, over the rendered observation $\hat{o}_i^{(t)}$ of \autoref{eq:obs_pv} in place of the entity sets of \autoref{eq:obs_vec}.
Transitions from all agents, acting through the shared policy $\pi_\theta$, are pooled into a single on-policy buffer, and training maximizes the standard clipped PPO surrogate~\cite{DBLP:journals/corr/SchulmanWDRK17} together with a value-regression loss for the critic and an entropy bonus.
Advantages and value targets use generalized advantage estimation~\cite{DBLP:journals/corr/SchulmanMLJA15}, the advantages normalized per minibatch and filtered following Gigaflow~\cite{gigaflow}.
\autoref{sec:appendix:training:opt} lists the full recipe and hyperparameters.

\paragraph{Policy architecture.}
\input{figs/pv_architecture}Inspired by DrivoR~\cite{drivor}, the perspective policy encodes each camera view $I_{i,k}^{(t)}$ of \autoref{eq:obs_pv} with a small convolutional stack, trained from scratch, and pools the resulting feature maps with a set of learned query tokens that cross-attend over them (\autoref{fig:pv_architecture}).
Richer than a single global reduction, this pooling also fixes the number of query tokens independently of the feature-map size, so cameras of differing resolution map into a latent of common width.
This branch replaces the entity-set encoders of the baseline vectorized policy from Gigaflow~\cite{gigaflow}; the pooled tokens are projected to the combined width of the embeddings they replace.
Everything else is inherited: the ego state $S$ and the conditioning $\mathcal{C}$ are encoded directly, and all embeddings are concatenated into a shared actor--critic MLP backbone with linear heads.
One further change, applied in both modalities, removes from $S$ two ego inputs a deployed vehicle cannot measure: the lateral offset from the lane center and the heading relative to the lane direction.
A performance gap between modalities (\autoref{sec:exp:driving}) therefore reflects the scene representation, not the architecture.
See \autoref{tab:arch-details} in \autoref{sec:appendix:arch} for the architectural details.

%% file: figs/system.tex
\begin{figure}[t]
  \centering
  \resizebox{\linewidth}{!}{%
  \includegraphics{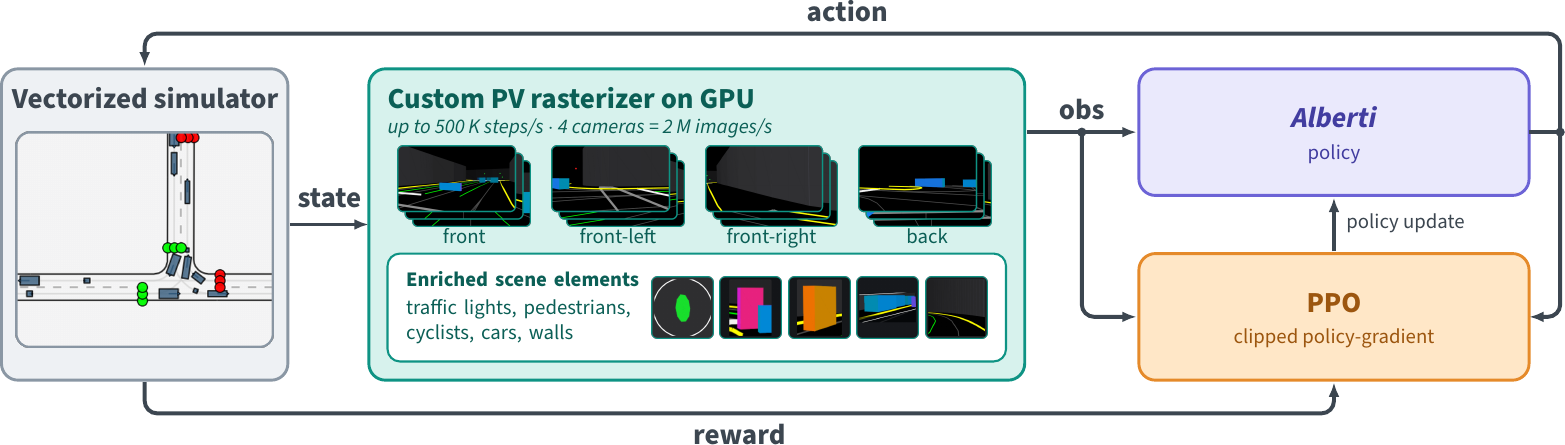}}
  \caption{\textbf{\ours architecture.}
    \ours couples a vectorized simulator with a custom GPU rasterizer that renders the simulator state into each agent's egocentric perspective view. The observation feeds \agent, the policy, and, together with the action and the reward, forms the rollout that PPO turns into a policy update.
    }
  \label{fig:system}
\end{figure}

%% file: figs/qualitative.tex
\begin{figure}[t]
  \centering
  \includegraphics[width=\linewidth]{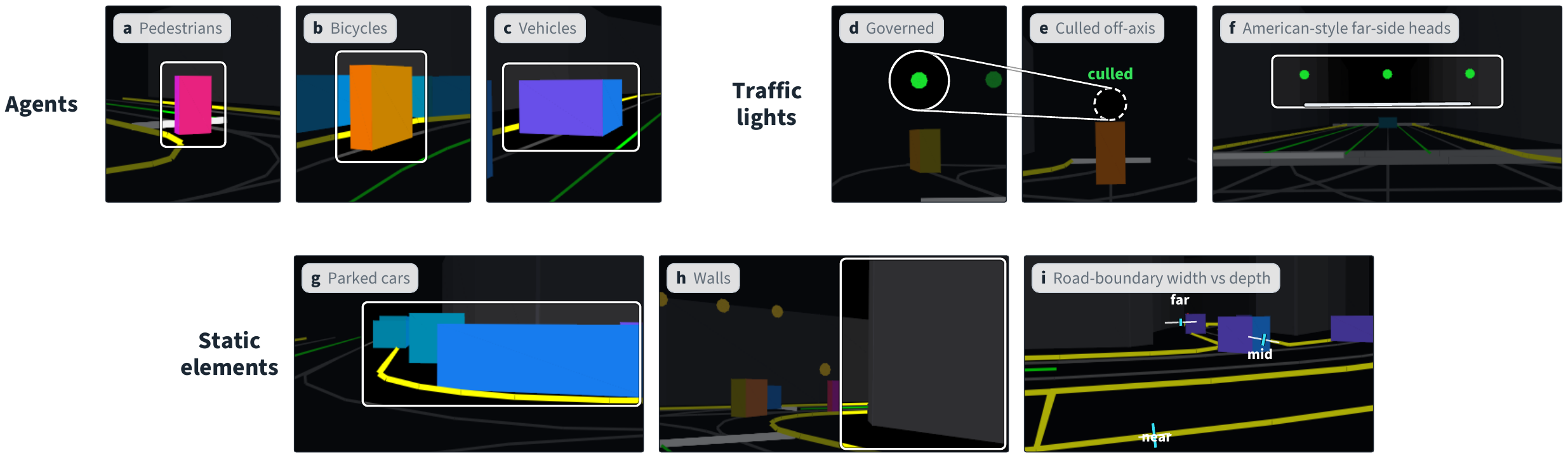}
  \caption{\textbf{Rendering features for driving.} Elements \ours draws into
    each agent's egocentric view, grouped by type.
    \emph{Agents} are cuboids colored per face with depth-aware brightness: 
    pedestrians~(a), cyclists~(b), vehicles~(c).
    \emph{Traffic lights} are disks with limited angular visibility, drawn only
    to the approach they govern~(d) and culled once the camera faces away~(e); the
    far-side heads of American-style intersections~(f).
    \emph{Static elements} are parked cars~(g), walls~(h), and road
    markings tapered with depth~(i).
    }
  \label{fig:qualitative}
\end{figure}

%% file: figs/pv_architecture.tex
\begin{wrapfigure}[18]{r}{0.44\linewidth}
  \centering
  \vspace{-2.5em}
  \resizebox{\linewidth}{!}{%
  \includegraphics{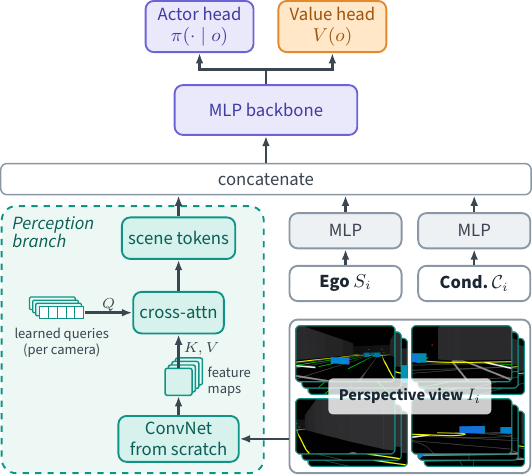}}
  \caption{\textbf{\agent architecture.}
    A from-scratch ConvNet encodes each camera view, and per-camera query tokens cross-attend over its feature map to pool it into scene tokens. The rest is unchanged from the vectorized agent.}
  \label{fig:pv_architecture}
\end{wrapfigure}

%% file: sections/04_rasterizer_eval.tex
\section{Rendering Efficiency Evaluation}
\label{sec:exp:renderer-analysis}
\ours puts perspective self-play at scale within reach thanks to its renderer. This section quantifies the two properties behind that role: its throughput exceeds that of prior renderers, and it accounts for only a small share of the training cost.

\input{figs/render_perf}

\paragraph{Throughput against prior renderers.}
\autoref{fig:render_throughput} (left) sweeps render resolution on a single A100L and reports peak render-only throughput in agent-steps/s, comparing \ours against the prior perspective renderers, each at the throughput reported in the Gigapixel paper~\cite{gigapixel}: RAP~\cite{rap}, a CPU rasterizer built on OpenCV; HUGSIM~\cite{hugsim}, a 3DGS renderer; and Gigapixel itself, the state-of-the-art batched renderer, built on the Madrona engine~\cite{madrona}.
RAP and HUGSIM sit orders of magnitude below the batched systems, so the contest that matters at self-play scale is with Gigapixel. Matching the square-resolution, wall-free configuration under which its numbers are reported, our CUDA rasterizer runs $1.4$--$4.1\times$ faster than its fastest configuration.
To isolate the engine from the scene content, we also reimplement our rasterizer in Madrona, so that both engines rasterize identical scenes. The reimplementation lands slightly below Gigapixel's published curve but follows the same trend across resolutions, indicating we exploit the engine about as effectively as Gigapixel's authors do.
All later comparisons use this reimplementation, not the Gigapixel simulator, so any difference reflects the rendering engine alone.

\paragraph{Scaling with hardware.}
Because \ours rasterizes on the general-purpose compute cores, it gets faster on newer hardware, where a conventional graphics-pipeline renderer does not. From
the A100L to the H100 (\autoref{fig:render_throughput}, right), our CUDA rasterizer speeds up by a uniform $1.5\times$; the Madrona reimplementation
instead \emph{slows}, to $0.2$--$0.7\times$ of its A100L throughput.
The margin widens from $2.2$--$5.0\times$ to $10$--$19\times$ across a single hardware generation.
The
cause is where each renderer does its work: ours runs on the compute cores and
inherits the H100's throughput gains, while Madrona runs on the fixed-function
graphics pipeline, which deep-learning accelerators deprioritize.
As accelerators keep trading graphics capacity for general-purpose compute, we expect this gap to widen further.

\paragraph{Self-play free of the renderer bottleneck.}
\autoref{fig:render_breakdown} breaks a training iteration into its phases, with each render covering a batch of $2048$ egos. Every non-render phase is identical for the two rasterizers, so the entire gap is in rendering --- $13$--$16\times$ slower on Madrona, and growing with resolution, while the CUDA rasterizer stays nearly flat. Rendering is therefore only about $10\%$ of a \ours iteration but over half of a Madrona one, making the full iteration $2$--$3\times$ slower; at the top resolution, $384\times216$, Madrona runs out of memory altogether. The env phase looks large only because this micro-benchmark runs serially, putting the CPU simulator on the critical path; real training overlaps it with compute, so rendering alone sets the throughput.
\autoref{tab:render_breakdown} in \autoref{sec:appendix:breakdown} gives the tabulated per-stage timings.
\autoref{sec:appendix:render} expands the full renderer comparison, from rendering quality to memory and policy quality per GPU-hour.

%% file: figs/render_perf.tex
\providecommand{\texttimes}{\ensuremath{\times}}
\ifdefined\rsw\else\newlength{\rsw}\fi
\ifdefined\rbw\else\newlength{\rbw}\fi
\begin{figure}[t]
    \centering
    \setlength{\rsw}{0.195\linewidth}%
    \setlength{\rbw}{0.195\linewidth}%
    \begin{subfigure}[t]{0.49\linewidth}
    \centering
    \includegraphics{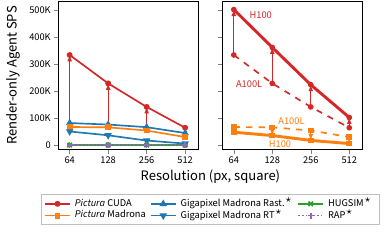}%
    \caption{Throughput comparison.}
    \label{fig:render_throughput}
    \end{subfigure}%
    \hfill
    \begin{subfigure}[t]{0.49\linewidth}
    \centering
    \includegraphics{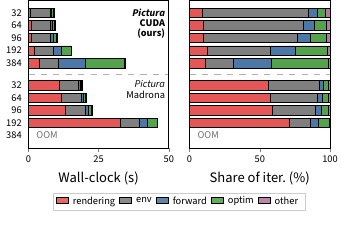}%
    \caption{Training-step wall-clock breakdown.}
    \label{fig:render_breakdown}
    \end{subfigure}
    \caption{\textbf{\ours renderer performance.}
    \textbf{(a)}~Render-only agent-steps/second vs.\ resolution. Left: %
    all systems on the same A100L; $^{\star}$numbers %
    from \cite{gigapixel}. Right: hardware scaling of the \ours CUDA renderer vs.\ its Madrona reimplementation, on A100L and H100.
    \textbf{(b)}~Per-phase wall-clock of a training iteration through render resolutions on a single H100 NVL, absolute seconds (left) and share of the iteration (right).
    }
    \label{fig:render_speed_compare}
\end{figure}

%% file: sections/05_policy_eval.tex
\section{Policy Evaluation}
\label{sec:exp}
This section evaluates \agent, \ours's perspective driving policy learned through large-scale self-play. \autoref{sec:exp:driving} presents the training and the driving performance it reaches, in domain and zero-shot on real-world-derived layouts, and \autoref{sec:exp:agent-analysis} analyzes how the policy grounds its driving decisions in what its cameras see.

\subsection{Driving Performance at Scale}
\label{sec:exp:driving}

\paragraph{Setup.}
We compare \agent against two vectorized baselines, matched in network size and number of collected state transitions: (1) the original privileged agent, which uses the same inputs as Gigaflow; and (2) a like-for-like variant from which we remove variables unavailable to the perspective agent---other-agent speeds, the ego vehicle's lateral offset from the lane center, and its angle relative to the lane direction. All three train on the CARLA maps (\autoref{sec:method:sim}), with \agent observing a four-camera rig rendered at $96\times54$ per camera. Driving quality is summarized by five vehicle-only metrics: at-fault collision rate (Coll.), off-road rate (Off-road), red-light violation rate (RLV), goals reached (Goals), and distance traveled per infraction (km/infr.). We evaluate two settings: (1) \emph{In-domain}, where agents are scored on newly generated self-play scenarios on the CARLA maps using all five metrics; and (2) \emph{Zero-shot}, where agents drive real-world-derived WOMD~\cite{womd} layouts re-rendered through \ours, with each vehicle assigned its logged start and final waypoint. In the zero-shot setting, we report the three applicable metrics: Coll., Off-road, and Goals. Because these layouts contain no traffic lights, RLV is undefined and km/infr.\ is not comparable to the in-domain setting, where red-light violations contribute to the infraction count. We evaluate three traffic densities in each setting: light, medium, and dense in-domain ($50$--$125\%$ of the maximum training density), and $50\%$, $75\%$, and $100\%$ of the recorded road users in WOMD.
See \autoref{sec:appendix:training} for the camera intrinsics and extrinsics, full training hyperparameters, reward-coefficient ranges, and hardware, and \autoref{sec:appendix:selfplay} for the evaluation protocols and metric definitions.

\input{tabs/selfplay_results}
\input{figs/alberti_training}

\paragraph{In-domain self-play evaluation.}
\autoref{tab:obs_comparison} reports in-domain self-play across three traffic densities. While prior large-scale self-play agents rely on privileged vector observations, we scale \agent's training to 50\,B agent steps using only rendered camera views. At medium traffic density, \agent nearly matches the like-for-like Vectorized$^\dagger$ reference in off-road rate ($0.018$ vs.\ $0.020$) and goals reached ($11.21$ vs.\ $12.81$), while exhibiting a moderately higher collision rate ($0.060$ vs.\ $0.037$). Its clearest gap lies in red-light violations ($0.058$ vs.\ $0.009$), which also largely account for its lower distance per infraction ($11.61$ vs.\ $26.99$\,km). This gap may arise because traffic lights span only a few pixels in the rendered views, whereas their exact status is directly provided in the vectorized state, even when they lie far ahead of the ego. The same overall pattern holds at light and dense traffic densities, showing that the policy generalizes to road-user densities different from that used during training. Despite having to learn visual perception jointly with driving, \agent uses the same training recipe and hyperparameters as the vectorized agent, without perception-specific tuning. \autoref{fig:alberti_training} further shows overall improvement across all five metrics with no clear sign of saturation by 50\,B steps, particularly for goals reached. These results suggest that both tailored optimization and further scaling could improve performance.

\paragraph{Zero-shot self-play on WOMD.}
On the zero-shot WOMD layouts (\autoref{tab:womd_selfplay}), the in-domain ranking inverts.
\agent now leads on all three metrics at full-density self-play, reaching $86\%$ of goals against the full privileged agent's $72\%$ and the like-for-like reference's $59\%$, with $2.5\times$ fewer collisions and $4.8\times$ less off-road than the full privileged agent.
This lead holds across traffic density, with \agent staying the safest and reaching the most goals at every level (100\%, 75\%, and 50\%) as each scene is thinned by removing a random fraction of its road users.
The full privileged agent still transfers better than the like-for-like reference, so the inversion is not about how much privileged information each carries.
It comes instead from the observation modality.
The rasterizer maps CARLA and WOMD geometry into the same visual vocabulary, so \agent's observations shift little between them, whereas the vectorized state exposes raw metric quantities such as lane geometry, spacing, and partner speeds, whose distributions differ between hand-authored and logged layouts.

\paragraph{Driving quality and throughput across resolutions.}
\autoref{fig:resolution} reports the driving performance of the \ours 10\,B agent at five render resolutions, alongside self-play throughput on $32$ H100s for a 10\,B training. Each resolution is a separately trained policy, so the panels show five discrete configurations. 
Safety metrics remain stable across resolutions, while goal completion and distance traveled per infraction improve consistently as resolution grows.
Throughput, shown in the rightmost panel, stays close to flat from $32\!\times\!18$ ($\sim$1.5\,M agent-steps/s) to the default $96\!\times\!54$ ($\sim$1.3\,M), then falls to $\sim$284\,K at $384\!\times\!216$, for a ${\sim}5\!\times$ drop overall.
Resolution buys goal completion and distance per infraction; throughput pays for it.
Because that cost sits at the top resolutions, raising resolution only near the end of training should reach high-resolution quality for a fraction of the compute.

\input{figs/resolution}

\subsection{Grounding in What the Cameras See}
\label{sec:exp:agent-analysis}
The metrics above aggregate over whole scenarios, averaging away the individual moments where the perspective-only \agent and the vectorized policy part ways. This analysis instead seeks out those moments---situations in which a difference in grounding would change what a policy does---asking whether the relevant information is represented and whether it is acted on. Grounding is what matters at deployment, where the perspective view is the only interface available.

\input{figs/value_removal}
\paragraph{Counterfactual probes reveal visual grounding.}
We first assess this grounding through two scalar quantities: the value $V(o)$ and the braking probability $P(\mathrm{brake}\mid o)$, the total probability assigned to all decelerating actions. For each nearby road user, we construct a counterfactual observation $o'$ by removing that user while leaving the rest of the scene unchanged. We then compute $\Delta V = V(o) - V(o')$ and $\Delta P(\mathrm{brake}) = P(\mathrm{brake}\mid o) - P(\mathrm{brake}\mid o')$, obtain the road user's occlusion percentage from the \ours renderer, and group the responses by occlusion level. Across 88\,K removals in 16 scenes, \autoref{fig:value_removal_agg} shows that \agent's $|\Delta V|$ and $|\Delta P(\mathrm{brake})|$ decrease as occlusion increases and drop to zero when the road user is fully hidden, whereas the privileged policy remains sensitive even then. The tracked pedestrian in \autoref{fig:value_removal_track} illustrates this contrast in a controlled setting. We move a pedestrian along a fixed path behind a vehicle; the strip below the curves indicates its visibility, from visible in green to occluded in red. As the pedestrian enters view, \agent's $\Delta V$ and $\Delta P(\mathrm{brake})$ depart from zero, with the braking response peaking near 20\,m, where the pedestrian is fully visible. Both responses then return toward zero when the pedestrian becomes occluded again. In contrast, the privileged policy remains sensitive even under occlusion: its $\Delta V$ stays away from zero, while its $\Delta P(\mathrm{brake})$ continues to vary substantially along the path. Thus, \agent's decisions depend on visual evidence, whereas the vectorized baseline can exploit scene state unavailable at deployment.

\input{figs/takeover_speed}
\paragraph{Closed-loop driving reveals occlusion-aware behavior.}
We further extend the grounding analysis in closed loop: we select recorded states in which the ego approaches a blind corner where an oncoming vehicle is occluded, and let each policy take over from the same state. All other agents continue along their recorded trajectories.
Across hundreds of paired takeovers (\autoref{fig:takeover_speed}\subref{fig:takeover_speed_agg}), \agent approaches the corner at a lower speed, resulting in a greater distance from the hidden vehicle when it becomes visible. From that point onward, the two policies decelerate at similar rates, showing that \agent establishes this safety margin before the vehicle becomes visible rather than through stronger deceleration afterward. At comparable blind corners with no occluded oncoming vehicle, the privileged agent accelerates, while \agent remains equally cautious.
In the example takeover (\autoref{fig:takeover_speed}\subref{fig:takeover_example}), \agent stops before the corner and proceeds cautiously until the vehicle becomes visible, whereas the privileged agent accelerates through the intersection ahead of it.
Camera-only training thus induces caution at the limits of visibility, whereas the privileged agent adjusts its speed using information unavailable to the deployed policy. This echoes the visibility asymmetry identified in LEAD~\cite{lead}: restricting the expert to sensor-observable evidence removes exactly this kind of non-causal demonstration and improves closed-loop performance. We therefore expect distillation from a privileged vector policy to inherit the same non-causal behavior unless the expert's observations are similarly restricted.

%% file: tabs/selfplay_results.tex
\begin{table}[t]
  \centering
  \caption{\textbf{Self-play driving performance, in-domain and zero-shot.}
    The same 50\,B-step recipe trained on three observation spaces. \textbf{(a)}~In-domain, on self-play in CARLA maps across
    three traffic densities. \textbf{(b)}~Zero-shot, on WOMD layouts, thinning traffic by removing a random fraction of road users per scene. All metrics are vehicle-only. \emph{Vectorized}$^{\dagger}$ drops the auxiliary features a camera cannot perceive.
    }
  \label{tab:selfplay_results}
  \vspace{-1.2em}
  \begin{subtable}[t]{0.57\linewidth}
    \centering
    \caption{In-domain self-play on CARLA maps.}
    \label{tab:obs_comparison}
    \resizebox{\ifdim\width>\linewidth\linewidth\else\width\fi}{!}{%
    \begin{tabular}{@{}llrrrrr@{}}
      \toprule 
      Density & Agent & Coll.\ \textdownarrow & Off-road \textdownarrow & RLV \textdownarrow & Goals \textuparrow & km/infr. \textuparrow \\
      \midrule
      \multirow{3}{*}{\begin{tabular}[c]{@{}l@{}}Dense \\ (125\%)\end{tabular}} & Vectorized & 0.049 & 0.024 & 0.008 & 12.61 & 23.46 \\
      & Vectorized$^{\dagger}$ & 0.038 & 0.019 & 0.008 & 10.32 & 22.25 \\
      & \agent & 0.063 & 0.018 & 0.043 & 8.94 & 11.44 
      \\
      \midrule
      \multirow{3}{*}{\begin{tabular}[c]{@{}l@{}}Medium\\ (100\%)\end{tabular}} & Vectorized & 0.043 & 0.027 & 0.009 & 15.38 & 33.53 \\
      & Vectorized$^{\dagger}$ & 0.037 & 0.020 & 0.009 & 12.81 & 26.99 \\
      & \agent & 0.060 & 0.018 & 0.058 & 11.21 & 11.61 \\

      \midrule
      \multirow{3}{*}{\begin{tabular}[c]{@{}l@{}}Light\\ (50\%)\end{tabular}} & Vectorized & 0.038 & 0.026 & 0.009 & 19.54 & 52.85 \\
      & Vectorized$^{\dagger}$ & 0.027 & 0.018 & 0.008 & 16.97 & 46.55 \\
      & \agent & 0.051 & 0.021 & 0.071 & 15.15 & 27.04 \\
      \bottomrule
    \end{tabular}%
    }
  \end{subtable}\hfill
  \begin{subtable}[t]{0.407\linewidth}
    \centering
    \caption{Zero-shot on WOMD.}
    \label{tab:womd_selfplay}
    \resizebox{\ifdim\width>\linewidth\linewidth\else\width\fi}{!}{%
    \begin{tabular}{@{}llrrr@{}}
      \toprule
      Density & Agent & Coll.\ \textdownarrow & Off-road \textdownarrow & Goals$^{\ast}$\textuparrow \\
      \midrule
      \multirow{3}{*}{\begin{tabular}[c]{@{}l@{}}100\%\\ (Full)\end{tabular}} & Vectorized           & 0.015 & 0.067 & 0.718\\
                           & Vectorized$^{\dagger}$ & 0.029 & 0.143 & 0.585\\
                           & \agent                 & 0.006 & 0.014 & 0.858\\
      \midrule
      \multirow{3}{*}{75\%} & Vectorized            & 0.009 & 0.070 & 0.746\\
                            & Vectorized$^{\dagger}$ & 0.022 & 0.144 & 0.607\\
                            & \agent                 & 0.003 & 0.013 & 0.870\\
      \midrule
      \multirow{3}{*}{50\%} & Vectorized             & 0.006 & 0.073 & 0.773\\
                            & Vectorized$^{\dagger}$ & 0.015 & 0.145 & 0.627\\
                            & \agent                 & 0.002 & 0.012 & 0.881\\
      \bottomrule
    \end{tabular}%
    }
    {\raggedright\tiny $^{\ast}$\,Only one goal is sampled per agent, so \emph{Goals} is a completion rate and cannot exceed~1.\par}
  \end{subtable}
\end{table}

%% file: figs/alberti_training.tex
\begin{figure*}[t]
\centering
\includegraphics{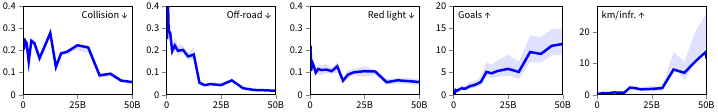}
\caption{\textbf{In-domain driving performance of \agent{} over 50\,B steps of self-play.} The bold curve is medium traffic density; the shaded band spans light to dense.}
\label{fig:alberti_training}
\end{figure*}

%% file: figs/resolution.tex
\begin{figure*}[t]
\centering
\includegraphics{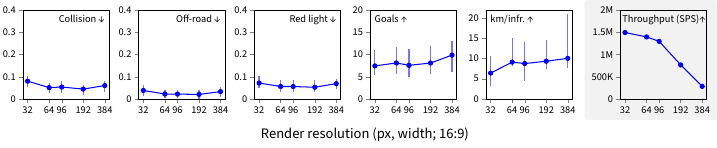}
\caption{\textbf{Rendering resolution sweep.} In-domain self-play driving metrics and training throughput of 10\,B-step \agent agents across render resolutions; a policy is trained for each resolution. The dot is the medium traffic density; the vertical lines spans the range across densities. Rightmost: peak self-play training throughput on 32 H100s.
}
\label{fig:resolution}
\end{figure*}

%% file: figs/value_removal.tex
\begin{figure}[t]
\centering
\begin{subfigure}[t]{0.34\linewidth}
\centering
\includegraphics{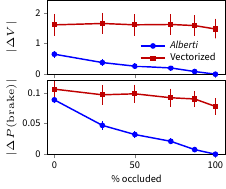}
\caption{Aggregate.}
\label{fig:value_removal_agg}
\end{subfigure}\hfill
\begin{subfigure}[t]{0.65\linewidth}
\centering
\begin{minipage}[t]{3.0cm}\centering
  \includegraphics[height=3.2cm,trim={0 0 3.5em 4em},clip]{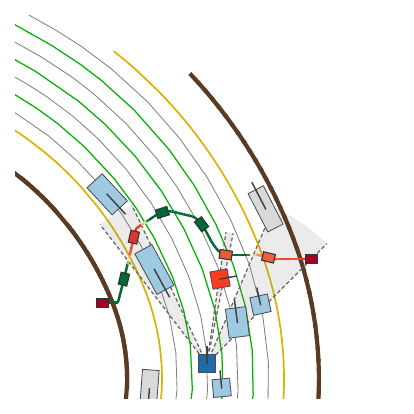}
\end{minipage}\hspace*{\fill}
\begin{minipage}[t]{4.7cm}\input{figs/track_stack}\end{minipage}%
\caption{One tracked pedestrian.}
\label{fig:value_removal_track}
\end{subfigure}

\caption{\textbf{Response of value and braking probability to a single nearby agent.}
  For \agent (blue) and the privileged \emph{Vectorized} baseline (red): the change in value $\Delta V$ and in braking probability $\Delta P(\mathrm{brake})$ caused
  by a single nearby road user.
  (\subref{fig:value_removal_agg})~\emph{Aggregate}: the magnitudes $|\Delta V|$ and
  $|\Delta P(\mathrm{brake})|$ from deleting one nearby agent, per-scene means against how occluded that agent is.
  (\subref{fig:value_removal_track}) One synthetic pedestrian swept along
  a path through a frozen scene: the top-down (left)
  draws the path colored by its visibility (green $\rightarrow$ red, fully visible to hidden), with the signed responses read along it (right).}
\label{fig:value_removal}
\end{figure}

%% file: figs/track_stack.tex
\includegraphics{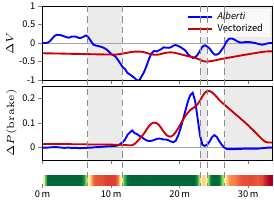}

%% file: figs/takeover_speed.tex
\begin{figure}[t]
\centering
\begin{subfigure}[b]{0.52\linewidth}
\centering
\sbox0{\includegraphics{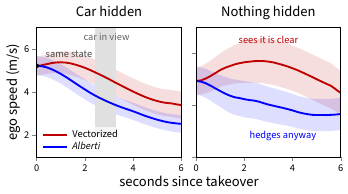}}\makebox[\dimexpr\wd0-8pt\relax][l]{\usebox0}
\caption{Aggregate over takeovers.}
\label{fig:takeover_speed_agg}
\end{subfigure}\hfill
\begin{subfigure}[b]{0.47\linewidth}
\centering
\includegraphics{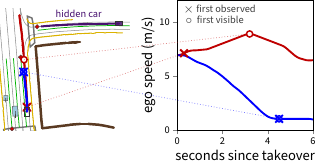}
\caption{One example takeover.}
\label{fig:takeover_example}
\end{subfigure}
\caption{\textbf{Closed-loop behavior at blind corners.} For the same recorded state, \agent (blue) and the privileged vectorized agent (red) take over the ego vehicle, while all other agents follow their logged trajectories. 
  (\subref{fig:takeover_speed_agg})~Mean ego speed with an occluded oncoming vehicle (left) and at comparable empty blind corners (right); the gray band marks the main reveal times.
  (\subref{fig:takeover_example})~Example takeover with cross traffic occluded by a building. $\times$: the hidden car first enters that agent's input; $\circ$: it first comes into line of sight. 
  }
\label{fig:takeover_speed}
\end{figure}

%% file: sections/06_conclusion.tex
\section{Conclusion}
\label{sec:conclusion}

Large-scale driving self-play does not require privileged observations: it can run directly on perspective images. \ours makes this practical with a custom CUDA rasterizer that renders each agent's view cheaply enough to train \agent from images under plain PPO. \agent is, to our knowledge, the first self-play driving policy trained from rendered camera views; it approaches the performance of its vectorized counterparts without access to their privileged observations, and transfers zero-shot to WOMD layouts. A gap remains between \ours's rasterized frames and real camera images. Closing it is orthogonal to the question we study here, and alignment methods~\cite{rap,gigapixel} can now start from a policy already grounded in what a camera sees. We release \agent{} as that starting point.

%% file: sections/0X_appendix.tex
\appendix

\makeatletter
\@ifundefined{theHchapter}{}{\renewcommand\theHchapter{app}}%
\renewcommand\theHsection{app.\arabic{section}}%
\renewcommand\theHsubsection{app.\arabic{section}.\arabic{subsection}}%
\makeatother

\section{Simulator and Training Details}
\label{sec:appendix:training}

This section details the per-agent reward and conditioning (\autoref{sec:appendix:reward}), the camera rig and policy architecture (\autoref{sec:appendix:arch}), and the optimization recipe, hyperparameters, and hardware (\autoref{sec:appendix:training:opt}).

\subsection{Reward and Conditioning}
\label{sec:appendix:reward}

The per-step reward is defined first, then the distributions from which the coefficients of the per-agent conditioning set $\mathcal{C}_i = (C_{\text{reward}}, C_{\text{dynamics}})$ of \autoref{sec:method:sim} are sampled.

\paragraph{Reward terms.}
The reward follows the Gigaflow design~\cite{gigaflow} and sums eleven terms,
\begin{equation}
\begin{split}
  R ={}& R_{\text{goal}} + R_{\text{collision}} + R_{\text{off-road}} + R_{\text{stop-line}} + R_{\text{comfort}} + R_{\text{l-align}} \\
       & + R_{\text{l-center}} + R_{\text{velocity}} + R_{\text{overspeed}} + R_{\text{reverse}} + R_{\text{timestep}}.
\end{split}
\label{eq:reward}
\end{equation}
Writing $p$ for the ego position, $g$ for the active goal, $v$ for the signed speed (m/s), $a_{\text{long}}, a_{\text{lat}}$ for the accelerations (m/s$^2$), $\dot{a}$ for the jerk, $\theta_f$ for the heading offset from the lane direction, $x_f$ for the signed lateral offset from the lane center (m), $d = |x_f - \alpha_{\text{center-bias}}|$, and $v_{\text{limit}}$ for the lane speed limit, the terms are
\begin{equation}
\begin{aligned}
  R_{\text{goal}} &= \mathbf{1}\big[\lVert p - g \rVert < \delta_{\text{goal}} \wedge (\text{waypoint} \vee v < v_{\text{goal}})\big] \\
  R_{\text{collision}} &= -\big(\alpha_{\text{collision}} + 0.1\,|v|\big)\, \mathbf{1}[\text{collision}] \\
  R_{\text{off-road}} &= -\alpha_{\text{off-road}}\, \mathbf{1}[\text{off-road}] \\
  R_{\text{stop-line}} &= -\alpha_{\text{stop-line}}\, \mathbf{1}[\text{red-light violation}] \\
  R_{\text{comfort}} &= -\alpha_{\text{comfort}} \big(\mathbf{1}[|a_{\text{long}}| > 3] + \mathbf{1}[|a_{\text{lat}}| > 3] \\
  &\qquad\qquad + \mathbf{1}[|\dot{a}_{\text{long}}| > 5 \vee |\dot{a}_{\text{lat}}| > 5]\big) \\
  R_{\text{l-align}} &= \alpha_{\text{l-align}}\, \Delta t\, \big(\!\min(\cos\theta_f, 0) + \alpha_{\text{vel-align}} \min(v \cos\theta_f, 0) \\
  &\qquad\qquad + 0.0025\, \big(1 - \tfrac{2}{\pi}|\theta_f|\big)\big) \\
  R_{\text{l-center}} &= -\alpha_{\text{l-center}}\, \Delta t\, \big(\mathbf{1}[\cos\theta_f > 0.5]\; d - 0.05\, e^{-(d - 0.5)}\big) \\
  R_{\text{velocity}} &= \alpha_{\text{velocity}}\, \Delta t\, \max(\cos\theta_f, 0)\, \mathbf{1}[v > 2.5] \\
  R_{\text{reverse}} &= -\alpha_{\text{reverse}}\, \Delta t\, \mathbf{1}[v < 0] \\
  R_{\text{overspeed}} &= -\alpha_{\text{overspeed}}\, \mathbf{1}[v > v_{\text{limit}} + 2] \\
  R_{\text{timestep}} &= -\alpha_{\text{timestep}}\, \Delta t\, \mathbf{1}[|v| > 0 \vee |a| > 0]
\end{aligned}
\label{eq:reward-terms}
\end{equation}
The overspeed penalty $R_{\text{overspeed}}$ is an addition relative to Gigaflow's reward: it fires when the speed exceeds the lane speed limit by more than 2\,m/s.
For pedestrian agents only the goal and collision terms are active: the remaining coefficients are zeroed and the goal radius is capped at 0.5\,m.

\paragraph{Conditioning distributions.}
Every coefficient of \autoref{eq:reward-terms} is resampled independently per agent at the start of each episode and exposed to the policy through the reward conditioning $C_{\text{reward}}$;
only $\alpha_{\text{velocity}}$ and $\alpha_{\text{timestep}}$ are kept fixed.
The dynamics conditioning $C_{\text{dynamics}} = (C_{\text{throttle}}, C_{\text{steer}}, C_{\text{acc}})$ collects randomized multipliers that scale the commanded longitudinal jerk, the commanded lateral jerk, and the maximum forward acceleration ($2.5\, C_{\text{acc}}$\,m/s$^2$), respectively.
\autoref{tab:reward-dist} lists all sampling distributions.

\input{tabs/reward_dist}

\subsection{Observation and Policy Architecture}
\label{sec:appendix:arch}

\paragraph{Camera rig.}
The perspective agent observes a four-camera rig $\mathcal{K}_i$ (front, front-left, front-right, and back) modeled on nuPlan's camera setup~\cite{nuplan}; \autoref{tab:camera_rig} lists its shared intrinsics and per-camera extrinsics.
The same rig is used for training and for evaluation.

\input{tabs/camera_rig}

\paragraph{Network.}
\autoref{tab:arch-details} lists the exact observation sizes and layer dimensions of the policies described in \autoref{sec:method:training}, at the default training configuration.
The trunk (backbone, heads, action space) is shared by both modalities;
the modalities differ only in the observation encoders.
The two vectorized baselines of \autoref{sec:exp:driving} share a single observation layout and network shape, and differ only in what fills it: the auxiliary slots a camera cannot perceive (ego lane-center distance, ego heading relative to the lane, and partner speed) carry their true values for \emph{Vectorized} and zeros for the like-for-like \emph{Vectorized}$^{\dagger}$.
The perspective observation carries no such slots.
Heavier pretrained image backbones (\eg, frozen DINO ViTs) are drop-in replacements behind the same token-pooling interface;
all reported results use the small from-scratch encoder.

\input{tabs/arch_details}

\subsection{Optimization and Hyperparameters}
\label{sec:appendix:training:opt}

\autoref{tab:hparams} lists the environment and optimization hyperparameters shared by all training runs of \autoref{sec:exp}.
All runs use the same single-stage PPO recipe, regardless of the observation modality.
Advantages are computed with GAE and normalized per minibatch.
Following Gigaflow's advantage filtering~\cite{gigaflow}, transitions whose advantage magnitude falls below $0.01$ times a running estimate of the maximum $|\hat{A}_t|$ (an exponential moving average with coefficient $0.25$ across rollouts) are dropped before the update; the surviving transitions are shuffled and reused for two epochs.
The learning rate follows a cosine schedule over the full run.
Multi-GPU runs split the agents and minibatches across ranks and keep the global sizes of \autoref{tab:hparams} unchanged.

\paragraph{Hardware.}
The perspective \agent runs of \autoref{sec:exp} use $32$ H100s ($8$ nodes of $4$), the configuration whose throughput \autoref{fig:resolution} reports; the vectorized baselines run on a single node ($4$ H100s). The render-resolution sweep of \autoref{fig:resolution} spans five 16:9 settings, $32\times18$, $64\times36$, $96\times54$ (the default), $192\times108$, and $384\times216$.

\input{tabs/hparams}

\section{CUDA Rasterizer vs. Madrona}
\label{sec:appendix:render}

This section compares three rasterizer configurations throughout: the \ours CUDA rasterizer, with the analytic-coverage antialiasing it runs in production; its Madrona reimplementation at Madrona's native one sample per pixel, the faster configuration the main text benchmarks; and a quality-matched antialiased Madrona baseline (AA) that recovers \ours's $2\times$ antialiasing by supersampling, rendering at doubled resolution and box-downsampling. Since every \ours number in the paper already includes its antialiasing, AA is the like-for-like quality comparison. The three are compared on rendering quality (\autoref{sec:appendix:render-quality}), on training-step time and thus on the wall-clock cost of matching that quality (\autoref{sec:appendix:breakdown}), on peak renderer memory (\autoref{sec:appendix:render-mem}), and finally on policy quality per GPU-hour (\autoref{sec:appendix:wallclock}), measured against Madrona in its fastest configuration.

\subsection{Rendering Quality}
\label{sec:appendix:render-quality}

\autoref{fig:town_scenes} places the three renders of each scene side by side. CUDA and single-sample Madrona are visually near-identical though not bit-exact; beyond antialiasing, Madrona also lacks per-face colors, so we encode agent orientation by coloring the whole box by whether the agent faces the ego.

The rows separate as resolution drops: the CUDA renders stay sharp, thin structures such as lane lines unbroken, while single-sample Madrona aliases into stair-stepped fragments below the training resolution and matches CUDA only as the image grows. Supersampling largely closes that gap, tracking the CUDA renders down to low resolutions, which is what makes it the like-for-like quality reference whose cost we measure in \autoref{sec:appendix:breakdown}. Matching \ours's quality in Madrona thus means rendering several times as many pixels, motivating the comparison that follows.

\input{figs/town_scenes}

\subsection{Training-Step Time Breakdown}
\label{sec:appendix:breakdown}

\autoref{fig:render_breakdown_aa} and \autoref{tab:render_breakdown} split a training iteration into its phases across the resolution sweep, for all three rasterizers; the main-text \autoref{fig:render_breakdown} shows the same breakdown without the AA baseline. The benchmark is a controlled single-H100-NVL micro-benchmark sized to one production rank: $256$ agents on $2$ environments, $128$-step rollouts, the $65{,}536$-transition minibatch with gradient accumulation ($32$ chunks of $2048$), the production policy ($64$-d group embeddings, $512$-wide backbone), compilation time excluded, and advantage filtering disabled so all three do identical forward and optimization work.

The env, forward, optimization, and other phases therefore agree to within measurement noise, and the entire wall-clock gap is rendering. Madrona rasterizes a $\max(W,H)^2$ square before cropping and pays a fixed per-call cost, so the many small render calls of gradient accumulation and the $128$-step rollout magnify its overhead and the slowdown widens with resolution; the CUDA rasterizer renders each camera at native size and scales gently with pixel count. Rendering is $13$ to $16\times$ slower on Madrona per iteration, and $14$ to $41\times$ on the AA baseline, which rasterizes four times the pixels. End to end the iteration is $2.1$ to $3.0\times$ slower, and up to $6.1\times$ with antialiasing, $58$ to $86\%$ of it spent rendering. \ours reaches $384\times216$, its lean render leaving room for the encoder forward, where both Madrona configurations run out of memory. Because the benchmark is serial the env phase sits on the critical path; production training overlaps it with GPU compute, so rendering governs throughput.

\input{figs/render_breakdown_aa}
\input{tabs/render_breakdown}

\subsection{Renderer Memory Usage}
\label{sec:appendix:render-mem}

\autoref{tab:render_mem} reports the peak GPU memory each rasterizer uses to render one $2048$-ego chunk at each resolution. It reports the device-level high-water mark (via a \texttt{cudaMemGetInfo} poller), which counts Madrona's external SimManager heap that PyTorch's allocator statistics miss, and isolates that fixed preallocated heap in a separate column. Madrona's heap depends only on the squared render size, so $192\times108$ and antialiased $96\times54$ (both a $192^2$ square) preallocate the same $13.2$\,GiB, and it grows steeply with resolution; the CUDA rasterizer preallocates nothing, streaming the batch in passes bounded to a fraction of free memory, so its peak stays much lighter (about $3\times$ less than Madrona at each resolution), and \ours and Madrona both reach $384\times216$ before running out of memory at $768\times432$. The three failures have distinct causes: antialiased Madrona is the first to fall over, at $384\times216$, where its heap and the downsample buffer together exceed the card; at $768\times432$ \ours runs out on the size of the output tensor alone, and Madrona on its preallocated heap alone.

\input{tabs/render_mem}

\subsection{Wall-Clock Quality per GPU-Hour}
\label{sec:appendix:wallclock}

A final comparison turns the throughput advantage into policy quality per GPU-hour. The same perspective agent is trained twice at $64\times36$, with the \ours CUDA renderer and its Madrona reimplementation at one sample per pixel, each capped at 2\,B agent steps, and \autoref{fig:wallclock_madrona} tracks the five vehicle-only metrics against wall-clock time. Madrona runs without antialiasing here, its fastest configuration and hence the most favorable to it, so the speedup reported below is a lower bound on the one against the quality-matched AA baseline. At matched steps both reach equivalent driving quality, so the renderer changes only how fast experience is collected, not each sample's worth. The result is a pure speedup: Madrona needs $3.35\times$ longer to reach the 2\,B-step cap ($10.2$ vs.\ $3.0$\,h), so at equal wall-clock the CUDA-rendered agent has seen proportionally more experience and leads on every metric.

\input{figs/wallclock_madrona}

\section{Policy Evaluation Details}
\label{sec:appendix:selfplay}

This section defines the driving metrics (\autoref{sec:appendix:metrics}) and the two evaluation settings (\autoref{sec:appendix:protocol}), and tracks the vectorized baselines over training (\autoref{sec:appendix:vec-baselines}).

\subsection{Metric Definitions}
\label{sec:appendix:metrics}

Driving quality is summarized by five vehicle-only metrics: at-fault collision rate (Coll.), off-road rate (Off-road), red-light violation rate (RLV), goals reached (Goals), and distance traveled per infraction (km/infr.).
Every rate is a per-episode indicator averaged over the controlled vehicles: an agent scores $1$ if the event occurs at any step of its episode and $0$ otherwise, so the reported rate is the fraction of vehicle episodes in which the event happens at least once. A collision is an oriented-bounding-box overlap between the ego and any other agent; it is charged as at-fault unless the ego is essentially stopped ($<\!0.2$\,m/s) or is struck from behind. Off-road fires when the vehicle box crosses a road-edge polyline, or its center leaves the drivable map. A red-light violation fires when the vehicle crosses, or changes lane into, a stop line held red by its controlling signal, within $10$\,m and with heading aligned to within $45^\circ$. A goal is reached when the vehicle comes within $2$\,m of it; Goals reports the mean number reached per vehicle, one per vehicle in the WOMD self-play, hence equivalently the fraction that arrive. Distance per infraction divides the total distance driven by the controlled vehicles (accumulated as speed\,$\times\,\Delta t$) by the number of vehicle episodes that contain at least one infraction, counting off-road, collision, and red-light events.

\subsection{Evaluation Protocol}
\label{sec:appendix:protocol}

Two settings are evaluated. \emph{In-domain}, each policy plays freshly generated self-play scenarios on the eight training maps at three fixed vehicle densities, light, medium, and dense ($60$, $120$, and $150$ agents per world, respectively 50\%, 100\%, and 125\% of the maximum number of agents per world seen during training); every density averages $200$ scenarios of $3000$ steps ($5$\,min), with actions taken deterministically (argmax). \emph{Zero-shot}, the same policies drive WOMD~\cite{womd} validation layouts re-rendered through \ours: every vehicle is controlled from its logged start toward its final logged waypoint, its single goal, and removed on arrival, while a vehicle already at its goal keeps log-replaying as context; episodes are capped at $600$ steps ($60$\,s), and $44{,}073$ of the $44{,}097$ validation scenes are scored (the other $24$ contain only parked vehicles). The in-domain setting reports all five metrics; the zero-shot setting reports the three relevant to it: Coll., Off-road, and Goals.

The two map sources also differ in how finely road geometry is stored, which the vectorized observation is sensitive to: the CARLA training maps are pre-chunked into segments of at most $10$\,m, whereas WOMD keeps raw polylines whose segments reach about $100$\,m. Each segment enters the observation as a single midpoint and half-length, so a coarse segment would stand in for a road edge running beside the ego with a point tens of meters away, and its length feature would leave the normalized range the policy was trained on. Long segments are therefore subdivided at observation time into equal pieces of at most a fixed span, which restores a training-like granularity. A sweep over that span puts the best zero-shot performance at $5$\,m, the value used for the reported numbers. The subdivision is applied to the WOMD layouts only; the in-domain evaluation runs on the maps' native $10$\,m chunking, unchanged from training. The perspective observation needs no such adjustment, since the rasterizer projects the polylines themselves.

\subsection{Vectorized Baselines over Training}
\label{sec:appendix:vec-baselines}

\input{figs/vec_training}

\autoref{fig:vec_training} tracks the two vectorized baselines over training, with and without the auxiliary features a perspective agent cannot perceive (ego lane-center distance, lane-heading, and partner speed). Both improve steadily and reach comparable driving quality by 50\,B agent steps, and the auxiliary-feature policy leads on lane-keeping (off-road and RLV), the terms its privileged inputs most directly inform. \agent's medium-density trajectory (dashed) tracks both baselines, closing on them as training proceeds, the perspective-view gap this paper studies (\autoref{tab:obs_comparison}).

%% file: tabs/reward_dist.tex
\begin{table}[t]
  \centering
  \caption{\textbf{Per-agent sampling distributions of the conditioning coefficients.}
    Reward coefficients (top, \autoref{eq:reward-terms}) and dynamics multipliers (bottom) are resampled per agent at every episode reset and exposed to the policy through the conditioning vector $C_i$. Each block is laid out as two coefficient/distribution column pairs, read left to right.
    The mixture $\mathcal{U}_{\text{mix}}(1/a, a)$ draws from $\mathcal{U}(1/a, 1)$ or $\mathcal{U}(1, a)$ with equal probability.}
  \label{tab:reward-dist}
  \setlength{\tabcolsep}{4pt}%
  \resizebox{\ifdim\width>\linewidth\linewidth\else\width\fi}{!}{%
  \begin{tabular}{@{}ll@{\hspace{2em}}ll@{}}
    \toprule
    Coefficient & Training distribution & Coefficient & Training distribution \\
    \midrule
    \multicolumn{4}{@{}l}{\emph{Reward conditioning $C_{\text{reward}}$}} \\
    $\delta_{\text{goal}}$ & $\mathcal{U}(2, 12)$ & $\alpha_{\text{vel-align}}$ & $\mathcal{U}(0, 1)$ \\
    $v_{\text{goal}}$ & $\mathcal{U}(0, 20)$ & $\alpha_{\text{l-center}}$ & $\mathcal{U}(2.5 \times 10^{-4}, 7.5 \times 10^{-3})$ \\
    $\alpha_{\text{collision}}$ & $\mathcal{U}(0, 3)$ & $\alpha_{\text{center-bias}}$ & $\mathcal{U}(-0.5, 0.5)$ \\
    $\alpha_{\text{off-road}}$ & $\mathcal{U}(0, 3)$ & $\alpha_{\text{velocity}}$ & $2.5 \times 10^{-3}$ (fixed) \\
    $\alpha_{\text{stop-line}}$ & $\mathcal{U}(0, 1)$ & $\alpha_{\text{reverse}}$ & $\mathcal{U}(2.5 \times 10^{-4}, 7.5 \times 10^{-3})$ \\
    $\alpha_{\text{comfort}}$ & $\mathcal{U}(0, 0.1)$ & $\alpha_{\text{overspeed}}$ & $\mathcal{U}(0, 1)$ \\
    $\alpha_{\text{l-align}}$ & $\mathcal{U}(2.5 \times 10^{-4}, 2.5 \times 10^{-2})$ & $\alpha_{\text{timestep}}$ & $2.5 \times 10^{-5}$ (fixed) \\
    \midrule
    \multicolumn{4}{@{}l}{\emph{Dynamics conditioning $C_{\text{dynamics}}$}} \\
    $C_{\text{throttle}}$ & $\mathcal{U}_{\text{mix}}(0.8, 1.25)$ & $C_{\text{acc}}$ & $\mathcal{U}_{\text{mix}}(2/3, 1.5)$ \\
    $C_{\text{steer}}$ & $\mathcal{U}_{\text{mix}}(0.8, 1.25)$ & & \\
    \bottomrule
  \end{tabular}}
\end{table}

%% file: tabs/camera_rig.tex
\begin{table}[ht]
  \centering
  \caption{\textbf{Camera rig.}
    The four-camera egocentric rig $\mathcal{K}_i$ modeled on nuPlan's cameras~\cite{nuplan}. Left: the intrinsics shared by all four cameras, with the render resolution used by default during training. Right: the per-camera extrinsics in the ego frame ($x$ forward, $y$ left, $z$ up, in meters; yaw about $z$). The three forward cameras overlap into a wide frontal field, and the back camera covers directly behind the vehicle.}
  \label{tab:camera_rig}
  \setlength{\tabcolsep}{4pt}%
  \resizebox{\ifdim\width>\linewidth\linewidth\else\width\fi}{!}{%
  \begin{tabular}[t]{@{}ll@{}}
    \toprule
    \multicolumn{2}{@{}l}{\emph{Shared intrinsics} (all cameras)} \\
    \midrule
    Sensor specification & $1920\times1080$ (16:9) \\
    Focal length         & $1545$\,px \\
    Field of view        & $63.7^\circ\times38.5^\circ$ \\
    Render resolution    & $96\times54$ \\
    \bottomrule
  \end{tabular}%
  \hspace{2em}%
  \begin{tabular}[t]{@{}lrrrr@{}}
    \toprule
    \multicolumn{5}{@{}l}{\emph{Per-camera extrinsics} (ego frame)} \\
    \midrule
    Camera & $x$ & $y$ & $z$ & yaw \\
    \midrule
    Front       & $1.66$  & $-0.01$ & $1.49$ & $0^\circ$   \\
    Front-left  & $1.63$  & $0.12$  & $1.48$ & $+55^\circ$ \\
    Front-right & $1.62$  & $-0.16$ & $1.49$ & $-55^\circ$ \\
    Back        & $-0.47$ & $0.02$  & $1.43$ & $180^\circ$ \\
    \bottomrule
  \end{tabular}}
\end{table}

%% file: tabs/arch_details.tex
\begin{table}[t]
  \centering
  \caption{\textbf{Policy architecture details.}
    Observation sizes and layer dimensions at the default training configuration.
    The perspective policy replaces the four entity-set encoders of the vectorized
    baseline with one image branch of exactly matching output width; the trunk is
    identical across modalities.
    The two vectorized baselines share one observation layout: the auxiliary slots are
    filled for \emph{Vectorized} and set to zero for \emph{Vectorized}$^{\dagger}$, so the
    network shape is the same for both.}
  \label{tab:arch-details}
  \begin{tabular}{@{}ll@{}}
    \toprule
    Component & Value \\
    \midrule
    \multicolumn{2}{@{}l}{\emph{Shared trunk (all modalities)}} \\
    Group embedding width & 64 \\
    Backbone & MLP, 4 layers $\times$ 512, GELU \\
    Actor, critic & shared backbone, linear heads \\
    Action space & joint discrete, 4 long.\ $\times$ 3 lat.\ jerk bins \\
    \midrule
    \multicolumn{2}{@{}l}{\emph{Vectorized observation encoders (both baselines)}} \\
    Ego vector & 7 floats (speed, size, steering, accel., speed limit) \\
    \quad auxiliary slots & $+2$ floats: lane-center distance, lane-heading cosine \\
    Partner features & 7 floats (rel.\ position, size, rel.\ heading) \\
    \quad auxiliary slot & $+1$ float: partner speed \\
    Entity sets & $\le$16 partners, $\le$80 lanes, $\le$80 boundaries, $\le$10 traffic ctrl. \\
    Per-group encoder & 2-layer MLP $\to$ 64-d, masked max over set members \\
    Conditioning & reward coefficients + 3 goal waypoints $\to$ 64-d \\
    Backbone input & $6 \times 64 = 384$-d \\
    \midrule
    \multicolumn{2}{@{}l}{\emph{Perspective observation encoder (replaces the entity sets)}} \\
    Cameras & 4 (front, front-left, front-right, back), $96\times54$\,px RGB (16:9) \\
    Conv stack & 5 layers, 128 output channels, trained from scratch \\
    Token pooling & 16 learned queries per camera, 1 cross-attention block \\
    Scene embedding & $4\times16$ tokens $\to$ 512-d ($=4\times128$) \\
    Ego + conditioning & same encoders as the baselines, without the auxiliary slots \\
    Backbone input & 384-d (identical) \\
    \bottomrule
  \end{tabular}
\end{table}

%% file: tabs/hparams.tex
\begin{table}[t]
  \centering
  \caption{\textbf{Training hyperparameters.}
    Self-play environment and PPO settings shared by all training runs.
    Rollout and minibatch sizes are in agent steps and are global: multi-GPU runs divide them evenly across ranks.}
  \label{tab:hparams}
  \begin{tabular}{@{}ll@{}}
    \toprule
    Parameter & Value \\
    \midrule
    \multicolumn{2}{@{}l}{\emph{Self-play environment}} \\
    Simulation timestep $\Delta t$ & 0.1\,s \\
    Episode length & 1280 steps (128\,s) \\
    Early episode reset & $\geq$40\% of agents inactive \\
    Training maps & 8 CARLA towns (Town01--Town07, Town10HD) \\
    Agents per map instance & $\sim \mathcal{U}\{1, 120\}$ \\
    Concurrent agents & $20 \text{ envs} \times 1024 = 20{,}480$ \\
    Pedestrian / cyclist fraction & 0.1 / 0.1 \\
    Parked cars per episode & 32 \\
    \midrule
    \multicolumn{2}{@{}l}{\emph{PPO optimization}} \\
    Rollout length & 128 steps \\
    Rollout size & $20{,}480 \times 128 = 2{,}621{,}440$ \\
    Minibatch size & $65{,}536$ \\
    Epochs per rollout & 2 \\
    Discount $\gamma$ & 0.999 \\
    GAE $\lambda$ & 0.95 \\
    PPO clip coefficient & 0.2 \\
    Value function clipping & none \\
    Value loss coefficient & 0.5 \\
    Entropy coefficient & 0.01 \\
    Optimizer & AdamW ($\beta_1{=}0.9$, $\beta_2{=}0.999$, $\epsilon{=}10^{-8}$, wd${=}0.01$) \\
    Learning rate & $5 \times 10^{-4}$, cosine schedule \\
    Max gradient norm & 0.5 \\
    Advantage normalization & per minibatch \\
    Advantage filtering threshold & $0.01 \times$ running max $|\hat{A}_t|$ \\
    Precision & FP32 \\
    \bottomrule
  \end{tabular}
\end{table}

%% file: figs/town_scenes.tex
\begin{figure}[t]
    \centering
    \includegraphics[width=\linewidth]{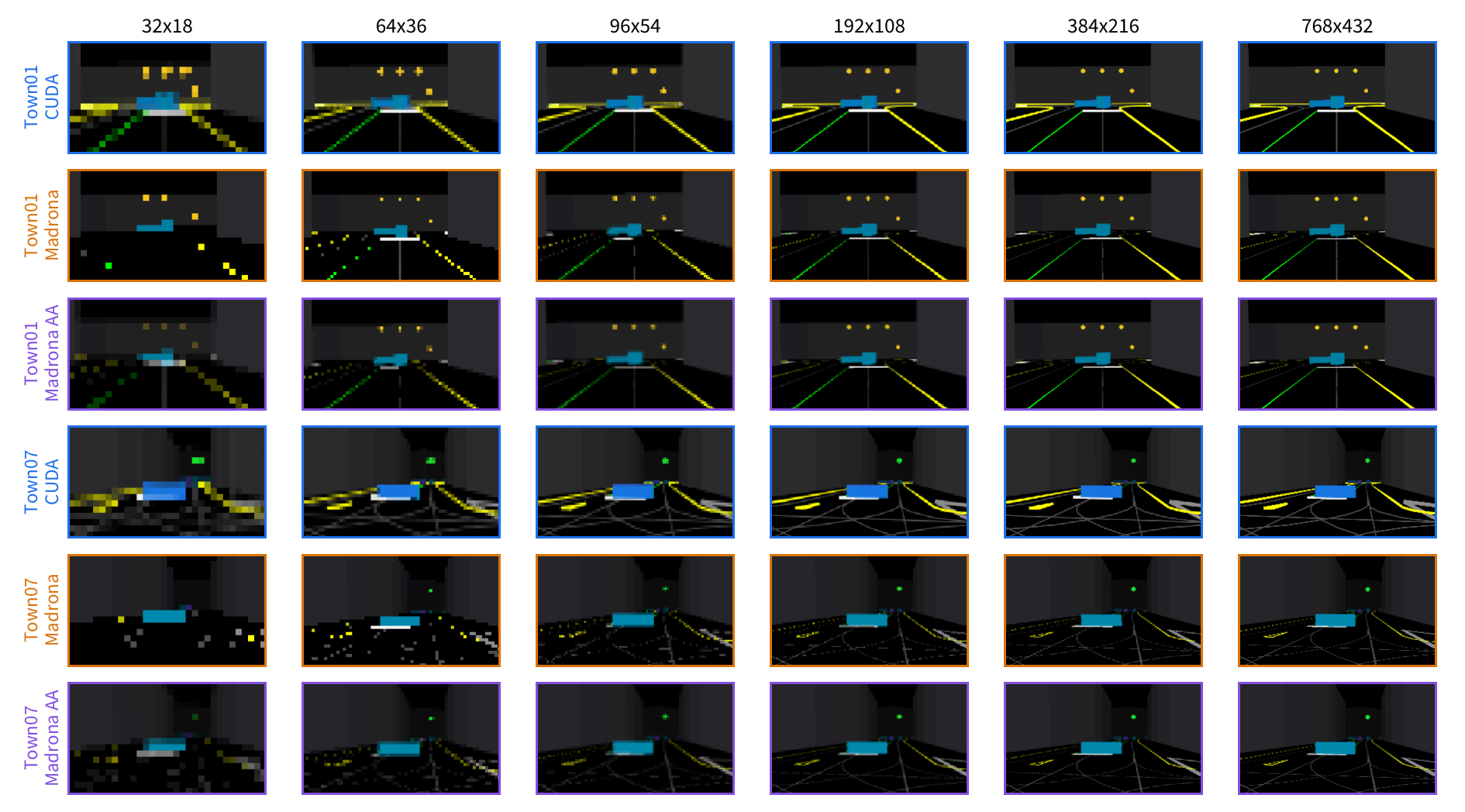}
    \caption{\textbf{Rendering across resolutions: CUDA vs.\ Madrona.}
      Front-camera renders of the same scenes. Each scene is shown in three rows: the \ours CUDA rasterizer (blue, analytic antialiasing), its Madrona reimplementation (orange, native one sample per pixel), and the quality-matched antialiased Madrona baseline (purple, $2\times$ supersampled) benchmarked in \autoref{fig:render_breakdown_aa}.}
    \label{fig:town_scenes}
\end{figure}

%% file: figs/render_breakdown_aa.tex
\providecommand{\texttimes}{\ensuremath{\times}}
\ifdefined\rbw\else\newlength{\rbw}\fi
\begin{figure*}[t]
  \centering
  \setlength{\rbw}{0.35\linewidth}%
    \includegraphics{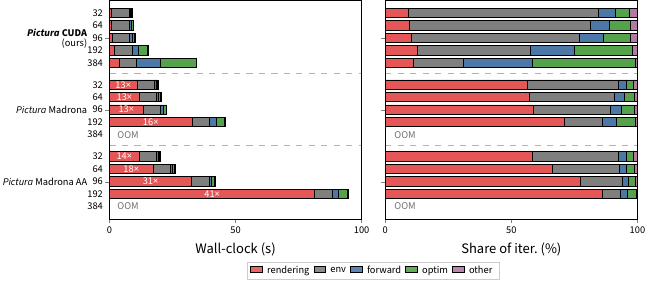}%
  \caption{\textbf{Training-step wall-clock breakdown including the quality-matched antialiased Madrona baseline.}
  The micro-benchmark of \autoref{fig:render_breakdown} extended with Madrona at $2\times$ supersampled antialiasing (AA), the quality-matched baseline against \ours's production $2\times$ AA; the \ours CUDA and Madrona rows repeat \autoref{fig:render_breakdown} for reference. White labels give each bar's rendering slowdown vs.\ \ours. Tabulated in \autoref{tab:render_breakdown}.}
  \label{fig:render_breakdown_aa}
\end{figure*}

%% file: tabs/render_breakdown.tex
\begin{table}[t]
  \centering
  \caption{\textbf{Per-phase training-step wall-clock: \ours CUDA rasterizer vs.\ Madrona reimplementation, without and with antialiasing.}
    Mean wall-clock per training iteration (seconds), split into env (the C simulator),
    rendering, policy forward, optimization, and other (host/device copies and bookkeeping),
    across the resolution sweep (single-H100-NVL micro-benchmark; setup in
    \autoref{sec:appendix:breakdown}). Peak render memory per rasterizer is in \autoref{tab:render_mem}.}
  \label{tab:render_breakdown}
  \resizebox{\ifdim\width>\linewidth\linewidth\else\width\fi}{!}{%
  \begin{tabular}{@{}llrrrrrrr@{}}
    \toprule
    Rasterizer & Resolution & env & rendering & forward & optim & other & Total & render $\times$ \\
    \midrule
    \ours (CUDA)     & $32{\times}18$   & 6.93 & 0.84  & 0.60 & 0.53 & 0.28 & 9.17  & --- \\
    Madrona          & $32{\times}18$   & 6.90 & 10.89 & 0.64 & 0.57 & 0.28 & 19.28 & $13.0\times$ \\
    Madrona (AA)     & $32{\times}18$   & 6.91 & 11.76 & 0.66 & 0.58 & 0.29 & 20.20 & $14.0\times$ \\
    \midrule
    \ours (CUDA)     & $64{\times}36$   & 6.90 & 0.94  & 0.75 & 0.80 & 0.27 & 9.65  & --- \\
    Madrona          & $64{\times}36$   & 6.97 & 11.82 & 0.78 & 0.80 & 0.28 & 20.65 & $12.6\times$ \\
    Madrona (AA)     & $64{\times}36$   & 6.92 & 17.25 & 0.78 & 0.82 & 0.28 & 26.05 & $18.4\times$ \\
    \midrule
    \ours (CUDA)     & $96{\times}54$   & 6.89 & 1.06  & 0.98 & 1.15 & 0.27 & 10.35 & --- \\
    Madrona          & $96{\times}54$   & 6.96 & 13.33 & 0.99 & 1.15 & 0.28 & 22.70 & $12.6\times$ \\
    Madrona (AA)     & $96{\times}54$   & 6.98 & 32.60 & 1.02 & 1.14 & 0.30 & 42.04 & $30.8\times$ \\
    \midrule
    \ours (CUDA)     & $192{\times}108$ & 6.93 & 1.99  & 2.68 & 3.57 & 0.28 & 15.45 & --- \\
    Madrona          & $192{\times}108$ & 6.94 & 32.70 & 2.62 & 3.52 & 0.32 & 46.11 & $16.4\times$ \\
    Madrona (AA)     & $192{\times}108$ & 6.99 & 81.39 & 2.58 & 3.39 & 0.31 & 94.66 & $40.8\times$ \\
    \midrule
    \ours (CUDA)     & $384{\times}216$ & 6.92 & 3.88  & 9.44 & 14.08 & 0.30 & 34.61 & --- \\
    Madrona          & $384{\times}216$ & \multicolumn{6}{c}{\emph{out of memory}} & --- \\
    Madrona (AA)     & $384{\times}216$ & \multicolumn{6}{c}{\emph{out of memory}} & --- \\
    \bottomrule
  \end{tabular}%
  }
\end{table}

%% file: tabs/render_mem.tex
\begin{table}[t]
  \centering
  \caption{\textbf{Peak GPU memory to render one training batch, across resolutions.}
    Peak device memory (GiB) to render a single $2048$-ego chunk ($8192$ scenes at $4$ views),
    measured in isolation on one $95$\,GB H100 NVL. \emph{Peak} is the device-level high-water
    mark, which counts Madrona's external SimManager heap that PyTorch's allocator statistics
    miss; \emph{heap} isolates that fixed, up-front allocation
    (\autoref{sec:appendix:render-mem}; benchmark setup in \autoref{sec:appendix:breakdown}).
    \emph{OOM} marks a configuration that exhausts the card.}
  \label{tab:render_mem}
  \setlength{\tabcolsep}{3.5pt}
  \begin{tabular}[t]{@{}llrr@{}}
    \toprule
    Res. & Rasterizer & Peak & Heap \\
    \midrule
    \multirow{3}{*}{$32{\times}18$} & \ours      & 5.6  & --- \\
                                    & Madrona    & 9.4  & 8.8 \\
                                    & Madrona AA & 10.3 & 9.2 \\
    \midrule
    \multirow{3}{*}{$64{\times}36$} & \ours      & 5.8  & --- \\
                                    & Madrona    & 10.5 & 9.2 \\
                                    & Madrona AA & 14.4 & 10.7 \\
    \midrule
    \multirow{3}{*}{$96{\times}54$} & \ours      & 6.3  & --- \\
                                    & Madrona    & 12.4 & 9.9 \\
                                    & Madrona AA & 21.1 & 13.2 \\
    \bottomrule
  \end{tabular}%
  \hspace{7mm}%
  \begin{tabular}[t]{@{}llrr@{}}
    \toprule
    Res. & Rasterizer & Peak & Heap \\
    \midrule
    \multirow{3}{*}{$192{\times}108$} & \ours      & 9.2  & --- \\
                                      & Madrona    & 22.6 & 13.2 \\
                                      & Madrona AA & 57.4 & 26.7 \\
    \midrule
    \multirow{3}{*}{$384{\times}216$} & \ours      & 20.5 & --- \\
                                      & Madrona    & 63.3 & 26.7 \\
                                      & Madrona AA & \multicolumn{2}{c}{\emph{OOM}} \\
    \midrule
    \multirow{2}{*}{$768{\times}432$} & \ours      & \multicolumn{2}{c}{\emph{OOM}} \\
                                      & Madrona    & \multicolumn{2}{c}{\emph{OOM}} \\
    \bottomrule
  \end{tabular}
\end{table}

%% file: figs/wallclock_madrona.tex
\begin{figure}[t]
\centering
\includegraphics{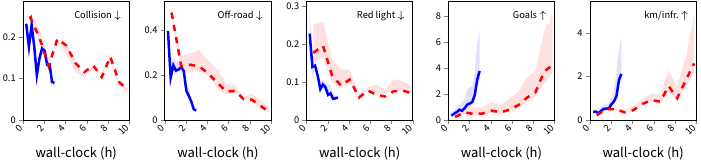}
\caption{\textbf{Wall-clock comparison: \ours{} (CUDA rasterizer) vs.\ the Madrona reimplementation.} The same perspective agent trained with either renderer at $64\times36$, both to 2\,B agent steps (\ours (CUDA), solid blue; Madrona, dashed red). Each panel tracks one vehicle-only metric, evaluated on held-out gigaflow self-play at medium density (120 agents, bold line; band spans light 60 to dense 150), against the wall-clock time at which each checkpoint was reached. The two renderers reach equivalent driving quality at matched steps, but Madrona's curves extend $3.35\times$ further in wall-clock ($10.2$ vs.\ $3.0$\,h to 2\,B steps). Wall-clock includes per-run startup and compilation.}
\label{fig:wallclock_madrona}
\end{figure}

%% file: figs/vec_training.tex
\begin{figure}[t]
\centering
\begin{subfigure}{\linewidth}
\centering
\includegraphics{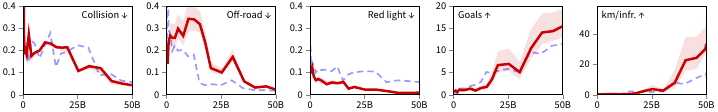}
\caption{Vectorized, \emph{with} auxiliary features.}
\label{fig:vec_ext_training}
\end{subfigure}

\vspace{0.35cm}

\begin{subfigure}{\linewidth}
\centering
\includegraphics{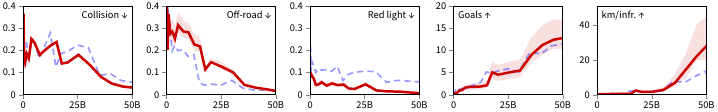}
\caption{Vectorized$^\dagger$, \emph{without} auxiliary features.}
\label{fig:vec_noext_training}
\end{subfigure}
\caption{\textbf{Vectorized self-play over training.} Self-play metrics of the two 50\,B vectorized policies across training, with and without the auxiliary features a perspective agent cannot perceive. Medium bold; band spans light--dense; the faint blue dashed curve is \agent's medium-density trajectory (\autoref{fig:alberti_training}).}
\label{fig:vec_training}
\end{figure}